\documentclass[10pt, a4paper]{article}

\usepackage{tabularx,booktabs,makecell,xcolor}
\usepackage{tcolorbox} 

\usepackage{multirow}
\usepackage{rotating}
\usepackage{tabularx}   
\usepackage{adjustbox}  
\usepackage{hyperref}

\usepackage{booktabs}

\usepackage[final]{lrec2026} 
\usepackage{pgfplots}
\pgfplotsset{compat=1.18}

\title{MedPT: A Massive Medical Question Answering Dataset for Brazilian-Portuguese Speakers\thanks{Accepted at LREC 2026.}}

\name{
\begin{tabular}{c}
Fernanda B. Färber$^{1,2}$, Iago A. Brito$^{1,2}$, Julia S. Dollis$^{1,2}$, Pedro S. F. B. Ribeiro$^{1,2}$ \\
Rafael T. Sousa$^{1,3}$, Arlindo R. Galvão Filho$^{1,2}$
\end{tabular}
}

\address{Advanced Knowledge Center for Immersive Technologies (AKCIT)$^1$\\Federal University of Goiás$^2$ \\ Federal University of Mato Grosso$^3$ \\
         \{fernandabufon, iagoalves, juliadollis, schindler\}@discente.ufg.br\\
         rafaelsousa@ufmt.br, arlindogalvao@ufg.br
         }

\abstract{
While large language models (LLMs) show transformative potential in healthcare, their development remains focused on high-resource languages. This creates a critical barrier for other languages, as simple translation fails to capture unique clinical and cultural nuances, such as endemic diseases. To address this, we introduce MedPT, the first large-scale, real-world corpus of patient-doctor interactions for the Brazilian Portuguese medical domain. Comprising 384,095 authentic question-answer pairs and covering over 3,200 distinct health-related conditions, the dataset was refined through a rigorous multi-stage curation protocol that employed a hybrid quantitative-qualitative analysis to filter noise and contextually enrich thousands of ambiguous queries, resulting in a corpus of approximately 57 million tokens. We further utilize of LLM-driven annotation to classify queries into seven semantic types to capture user intent. To validate MedPT's utility, we benchmark it in a medical specialty classification task: fine-tuning a 1.7B parameter model achieves an outstanding 94\% F1-score on a 20-class setup. Furthermore, our qualitative error analysis shows misclassifications are not random but reflect genuine clinical ambiguities (e.g., between comorbid conditions), proving the dataset's deep semantic richness. We publicly release MedPT on \href{https://huggingface.co/datasets/AKCIT/MedPT}{Hugging Face} to support the development of more equitable, accurate, and culturally-aware medical technologies for the Portuguese-speaking world.
 \\ \newline  }

\begin{document}

\maketitleabstract

\section{Introduction}

Large language models (LLMs) are rapidly expanding their reach across diverse healthcare applications, from online patient support and clinical text classification to robust medical question answering \cite{nazi2024large, 10819409}. However, the performance, safety, and clinical reliability of these systems are fundamentally contingent on the scale and quality of their training data \cite{10.1007/978-3-031-40923-3_16}. Consequently, datasets that lack demographic representation, contain inherent biases, or are insufficiently curated to handle complex medical nuances can lead to models that produce factually incorrect or systematically skewed outputs, posing significant risks and undermining clinical trust \cite{KHAN2021130, schwabe2024metric}. Compounding this challenge, current progress and resource development are overwhelmingly concentrated in high-resource languages, particularly English, a linguistic bias that creates a critical barrier for other major world languages, severely limiting the development, validation, and equitable deployment of these promising technologies to large global populations \cite{10620223}.

This linguistic and data-centric gap is clearly reflected in the literature, where significant research investment in medical text curation has centered on English \cite{jin2019pubmedqa, pmlr-v174-pal22a, singhal2022large, preston2023toward}, Chinese \cite{he2019applying, 8548603, ZHU2023102573}, or both \cite{he2020meddialog}. Despite valuable initiatives have begun to extend this scope to other languages such as Arabic, Vietnamese, Persian, and Turkish \cite{abdelhay2023deep, nguyen2022spbertqa, ghassabi2025leveraging, 10711128}, the resource disparity remains particularly acute for Brazilian Portuguese in specialized domains, such as medical question answering.

While translating existing datasets may seem like a viable shortcut, this approach is fundamentally limited because it fails to capture essential local context. For example, the epidemiology of endemic conditions such as Dengue and Chagas disease is entirely absent from these resources. Furthermore, translation erases crucial linguistic and cultural nuances, including region-specific clinical expressions and colloquial patient descriptions of symptoms \cite{naveen2024overview}. This creates a dual problem: locally, it constrains the advancement of clinically reliable language technologies for the population. Globally, it perpetuates a critical void in the medical data landscape, which remains heavily biased towards the health contexts prevalent in high-resource language communities. Therefore, creating a robust dataset for Brazilian Portuguese is not only essential for its speakers but also represents an opportunity to enrich global health knowledge with data on otherwise underrepresented diseases and clinical realities.

In this work, we introduce MedPT\footnote{\href{https://huggingface.co/datasets/AKCIT/MedPT}{https://huggingface.co/datasets/AKCIT/MedPT}}, the first public large-scale corpus for the Brazilian Portuguese medical domain, comprising 384,095 authentic question-answer pairs from real patient-doctor interactions along with metadata detailing the disease, the medical specialty, and a label identifying the question type. We provide a comprehensive analysis of the dataset, detailing its linguistic properties, such as the natural asymmetry of patient and doctor communication, and its thematic breadth, which covers over 3,200 medical conditions across 104 medical specialties. To validate its practical utility, we establish strong benchmark on a real-world medical speciality classification task, achieving 94\% F1-score on a challenging 20-class setup by fine-tuning a modern 1.7B parameter model on MedPT, confirming the dataset is a high-quality and effective resource for developing and specializing language models for the nuanced Brazilian healthcare context.

This resource provides a foundational dataset for a range of tasks, including medical question answering, patient intent classification, and linguistic studies on doctor-patient communication. More broadly, MedPT captures the real-world nuances of healthcare interactions, enabling the development of more contextualized, reliable, and accessible models for Portuguese-speaking populations.

\section{Related Works}

In recent years, several datasets have been developed to support research in medical question answering and health-related natural language processing (NLP). These resources have been fundamental for training models capable of reasoning over clinical information, understanding patient intent, and generating reliable medical responses. Natural language processing has been applied across multiple medical fields, including oncology, psychiatry, and primary care \cite{cascella2024breakthrough, wang2020systematic, 10.1093/jamia/ocz200}. Within these domains, the most common NLP tasks are information extraction \cite{10.1093/jamia/ocab126, cury2021natural, 8787232}, classification \cite{magna2020application}, named entity recognition \cite{ji2019hybrid} and question answering \cite{singhal2025toward}.

However, the development and fine-tuning of such models require large, well-structured, and high-quality datasets. To address this, several datasets have been proposed to advance research in medical natural language processing. For example, \citet{johnson2023mimic} introduced MIMIC-IV, which focuses on describing a vast collection of de-identified electronic health records from intensive care units, widely used for developing clinical prediction models and training NLP systems on real-world clinical documentation. Similarly, BioRED \cite{10.1093/bib/bbac282} aimed at advancing biomedical knowledge extraction by annotating document-level relations between diverse entities like genes, diseases, and chemicals from scientific papers abstracts.

Despite these contributions, most existing datasets consist primarily of clinical or biomedical records and lack open-ended, real-world patient questions paired with verified medical answers. \citet{he2019applying} proposed cMedQA, which includes 54,000 pairs of physician-patient questions and answers in Chinese, sourced from online health forums, focusing on non-restricted, open-ended questions. Similarly, MedRedQA \cite{nguyen2023medredqa} includes 51,000 pairs of open-ended English consumer questions and verified expert answers, collected from online medical consultation forums.

Moreover, pioneering efforts have begun to establish benchmark medical datasets for several under-resourced languages, employing a variety of curation strategies. One common approach involves collecting data from trusted online health portals. For instance, ViHealthQA \cite{nguyen2022spbertqa} contains over 10,000 medical question-answering pairs for Vietnamese sourced from credible health websites, while MAQA \cite{abdelhay2023deep} represents the largest Arabic healthcare Q\&A dataset, with over 430,000 questions from regional health platform. Other initiatives have utilized different sources; MedTurkQuAD \cite{10711128}, a Turkish medical QA dataset, was derived from Turkish Wikipedia and national medical theses. The effort for Persian by \citet{ghassabi2025leveraging} was even more comprehensive, introducing the MF3QA dataset and leveraging it to create Gaokerena-V1, the first open-source Persian medical language model. Together, these works illustrate a growing cross-lingual interest in expanding medical NLP resources beyond English, particularly for morphologically rich or culturally distinct languages.

Recent efforts to address the Portuguese data shortage include the work of \citet{paiola2024adapting}, who fine-tuned LLMs on an English healthcare corpus translated into Brazilian Portuguese. While this approach enabled preliminary evaluations of medical dialogue assistants, the resulting dataset was not made publicly available. More fundamentally, its reliance on translation means it inherently lacks the authentic, region-specific linguistic nuances and clinical contexts, such as endemic diseases, that only a native, real-world dataset can capture. This highlights a persistent and critical need for a publicly accessible, native corpus built from authentic patient-doctor interactions to advance the field.

Alongside translation efforts, existing medical datasets for native Brazilian Portuguese have centered on formal clinical text from hospital records. Prominent examples include BRATECA \cite{consoli2022brateca}, a large collection of de-identified clinical notes; SemClinBr \cite{oliveira2022semclinbr}, a semantically annotated corpus for entity recognition; and the work of \cite{da2019contributions}, identifying symptoms in hospital texts. While valuable, these resources are composed of formal medical documentation and do not capture the open-ended, conversational dialogues between patients and physicians. This absence of real-world, patient-initiated questions is a primary barrier to developing and evaluating robust interactive healthcare systems in Portuguese.

A recent initiative targeting this conversational gap in Portuguese is IaraMed \cite{farber2025iaramed}, a women’s healthcare chatbot developed using data scraped from online health platform. While a promising application of real-world medical Q\&A, this work is narrowly focused on women's health and does not cover the full spectrum of patient-doctor inquiries, and the underlying dataset was not publicly released. Thus, a critical need persists for a broad-domain, publicly accessible, native corpus built from authentic patient-doctor interactions to advance the field.

\section{MedPT}

This section details the multi-stage process applied in the development of MedPT dataset, designed to transform raw, user-generated content into a robust language resource. We first describe the data acquisition from an authentic patient-doctor interaction platform, followed by our extensive data cleaning and curation pipeline, which includes both general preprocessing and nuanced, column-specific refinements. We then detail an LLM-driven annotation step, where we employed a large language model to classify each user query and identify its question type (e.g., treatment, diagnosis). to enhance the dataset’s utility. Finally, to ensure a fair comparison and compatibility with modern models, all metrics were extracted using the sub-word tokenizer adopted by state-of-the-art LLMs \cite{achiam2023gpt}.

\subsection{Data Acquisition}
To build the MedPT dataset, we collected real medical questions and answers in Portuguese from Doctoralia\footnote{https://www.doctoralia.com.br}, a widely used online platform where Brazilian patients ask health-related questions and receive responses from verified doctors. Doctoralia was chosen as data source due both its large volume of questions across a wide range of medical specialties and its verification process to ensure that only certified health professionals can respond to users’ inquiries. Additionally, all questions are completely anonymous, which guarantees the privacy and confidentiality of users’ personal information. Furthermore, the data collected from this platform are natural and user-generated, reflecting real-life inquiries from Brazilian users. This aspect makes the dataset particularly valuable for developing healthcare solutions tailored to the Brazilian context, as it captures the country’s linguistic and cultural nuances, healthcare system characteristics, and the prevalence of region-specific diseases.

To collect the questions, answers, and related data, we developed a custom web-scraping pipeline to extract information directly from the website. Each sample includes the medical condition of the question (e.g., hypertension, pregnancy, diabetes), the patient’s textual query, the doctor’s textual response, and metadata about the physician, such as specialty and rating. During this collection phase, we gathered approximately 2 million raw samples.

\subsection{Data Cleaning}
The initial cleaning phase focused on structural normalization and deduplication. A primary source of data redundancy was identified in the platform's design, which allowed a single question-answer pair to be associated with multiple medical specialties. This resulted in numerous duplicate entries for the same core text. To address this and consolidate the dataset, we removed the redundant specialty column and then applied a strict deduplication process based on the core question-answer text. This procedure, combined with the removal of other irrelevant columns (e.g., doctor's ratings), reduced the corpus from approximately 2 million raw samples to 393,143 unique entries. Subsequently, we performed a textual normalization step across all columns, removing HTML tags, URLs, special characters, and extraneous whitespace, which significantly reduced noise as evidenced by a decrease in the average token count per question-answer pair from 33.05 to 26.54 tokens.

\subsubsection{Column-Specific Curation}
To ensure the quality and suitability of the corpus for question-answering tasks, we implemented a multi-stage, column-specific curation protocol through all features of our dataset (\textit{Question}, \textit{Answer} \textit{Condition} and \textit{Medical Specialty}). This protocol combined quantitative analysis with qualitative manual inspection. We first analyzed token distributions such as mean, standard deviation (SD) and range to identify statistical outliers, which were then manually reviewed to confirm their nature. Based on this hybrid analysis, we applied targeted procedures, including the removal of malformed or uninformative entries and the contextual enrichment of incomplete questions, as detailed below.

\paragraph{Question.} The \textit{Question} column exhibited high linguistic diversity (mean 42.5, SD 31.0), with token counts ranging from 3 to 801. Our qualitative inspection confirmed that the longest entries were valid, complex patient narratives. Conversely, the shortest entries (3–9 tokens) were typically found to be context-dependent (e.g., "What is the treatment?"), lacking sufficient information to be standalone queries. To address this ambiguity without data loss, we implemented a contextual enrichment procedure: the corresponding \textit{Condition} was prepended to each of these 7,178 incomplete queries. This process rendered each question self-contained and preserved a significant portion of the dataset that would have otherwise been discarded.

\paragraph{Answer.} The \textit{Answer} column exhibited even greater variability (mean 96.5, SD 75.0), with token counts spanning from 0 to 2,984. To ensure the corpus contained only direct, high-quality responses, a qualitative review of these extremes was performed. This analysis revealed two distinct types of low-utility data: answers with fewer than 10 tokens were found to be uninformative (e.g., "Consult a specialist"), while those exceeding 1,000 tokens, though medically valid, were typically generic informational articles rather than specific replies to the patient's query. To optimize the dataset for direct question-answering tasks, we established a valid token range of 10–1,000. Applying this filter removed 8,359 samples, significantly improving the signal-to-noise ratio and focus of the final corpus.

\paragraph{Condition.}
For the \textit{Condition} column, we implemented a token-count-based filtering protocol to ensure data quality. An initial analysis programmatically removed all zero-token samples. We then conducted a qualitative review of the distribution's extremes. This inspection revealed that queries containing very few tokens (e.g., "anxiety") were valid medical conditions and were therefore retained. Conversely, entries with the highest token counts were found to be consistently malformed or corrupted data. Based on this manual review, we established a conservative upper threshold of 23 tokens to precisely exclude these erroneous entries while preserving all valid conditions. This procedure refined the column's consistency by removing 398 samples.

\paragraph{Medical Specialty.} The Medical Specialty column presented a distinct curation challenge. Statistical analysis showed moderate variance (mean 5.7 tokens, SD 3.4), which our qualitative review attributed to the platform's feature allowing professionals to list multiple specialties. We made the deliberate decision to retain these multi-specialty entries to avoid information loss, as they represent valid, rich data. However, a separate filtering process was necessary to address genuine scraping artifacts. Our analysis identified and removed non-specialty data, such as location names (e.g., "Fortaleza"), numerical values, and empty strings. This targeted procedure refined the column's integrity by excluding 62 invalid samples.

Following this multi-stage cleaning protocol, we achieve a final dataset comprising 384,095 high-quality question-answer pairs, resulting in a substantial and reliable corpus tailored for developing language models in the Brazilian Portuguese medical domain.

\subsection{Question-Type Annotation}\label{synthetic}
\label{sec:annotation}

To provide a fine-grained, semantic understanding of user intent within the corpus, we enriched the dataset with a synthetically generated feature column classifying the type of each user question. This structural annotation is critical for analyzing the dataset's composition (e.g., quantifying the types of user needs) and enables more targeted modeling applications. Following established practices in dataset construction \cite{alhuzali2024mentalqa, guo2018qcorp}, we employed gpt-oss-120b \cite{agarwal2025gpt} to annotate each user query. Through a structured prompt, avalable in Appendix~\ref{app:prompt}, the model assigned each question to one of the following seven predefined classes, which represent distinct facets of health-related inquiries:

\begin{itemize}
\item \textbf{Anatomy and Physiology:} Groups questions about the structure and function of body organs and systems.
\item \textbf{Choosing Healthcare Professionals:} Captures doubts about which specialist to consult or how to access medical services.
\item \textbf{Diagnosis:} Questions focused on the interpretation of symptoms, medical exams, or the identification of diseases.
\item \textbf{Epidemiology:} Covers topics such as causes, risk factors, and disease complications.
\item \textbf{Healthy Lifestyle:} Refers to questions about habits that influence health, such as diet, exercise, or sleep.
\item \textbf{Treatment:} Encompasses questions about medications, therapies, side effects, and contraindications.
\item \textbf{Other:} Includes questions that do not fit into the previous categories but are still related to health or the healthcare system.
\end{itemize}

\section{Dataset Characterization}
To characterize the final MedPT corpus and validate its suitability for training robust language models, we performed an extensive exploratory data analysis. This section details the dataset's key statistical and structural properties, including an analysis of the distribution across question types and medical specialties, a summary of its linguistic features, and a presentation of representative samples. The analysis underscores the corpus's large scale and multi-domain nature, confirming its value as a comprehensive resource for the Brazilian Portuguese medical NLP community.

\subsection{Dataset Overview}
The MedPT dataset is a large-scale corpus for medical question answering in Brazilian Portuguese. Each entry comprises a patient's question paired with one or more verified answers from licensed healthcare professionals, ensuring both authenticity and clinical reliability. The dataset is structured with rich metadata to support a diverse range of downstream NLP tasks. The fields are detailed as follows:

\begin{itemize}
\item \textbf{Question:} The patient's original inquiry in Brazilian Portuguese, typically describing symptoms, test results, or seeking information about medications and treatments.
\item \textbf{Answer:} A response provided by a licensed healthcare professional, containing clinically grounded explanations or recommendations.
\item \textbf{Condition:} A finer-grained topic within the main field, often a specific disease or condition (e.g., \textit{Diabetes}, \textit{Acne}, \textit{Hypertension}).
\item \textbf{Medical Specialty:} The specialization of the professional who authored the answer (e.g., \textit{General Practitioner}, \textit{Psychologist}, \textit{Nutritionist}).
\item \textbf{Question Type:} The semantic intent of the question, classified into one of seven categories: \textit{Diagnosis}, \textit{Treatment}, \textit{Anatomy and Physiology}, \textit{Epidemiology}, \textit{Healthy Lifestyle}, \textit{Choosing Healthcare Professionals}, or \textit{Other}.
\end{itemize}

Collectively, these fields form a richly structured dataset comprising over 384K question-answer pairs, derived from more than 180K unique questions. This composition enables a wide range of applications, from specialty prediction and question-type classification to the development and domain adaptation of medical language models.

\subsection{Quantitative Analysis}
The scale and structure of MedPT are defined by its core statistics. A key characteristic of the dataset is its one-to-many mapping, with each question receiving an average of 2.11 answers. While most questions receive between one and three responses, some attract a much larger number, with an observed maximum of 266 answers for the same question. This multiplicity is a central feature of the corpus, reflecting the platform’s collaborative nature where different healthcare professionals provide complementary clinical perspectives on a single patient inquiry.

To characterize the linguistic properties of the corpus, we performed a quantitative analysis of its primary textual fields. Together, the question and answer columns contain nearly 57 million tokens distributed in a natural asymmetry that reflects the platform's core pragmatic context. Patient questions are typically concise while the corresponding professional answers are substantially more detailed and explanatory (averaging 51.79 and 96.49 tokens length, respectively). This divergence stems from the distinct communicative goals of each role: patients are focused on symptom description and problem specification, whereas doctors are tasked with providing comprehensive explanations, differential considerations, and essential risk-management disclaimers.

\begin{table}[t]
\small
\centering
\renewcommand{\arraystretch}{1.2} 

\adjustbox{width=0.48\textwidth}{
\begin{tabular}{lcc}
\toprule
\textbf{Condition} & \textbf{Questions} & \textbf{Answers} \\
\midrule
\textbf{Frequent - Questions}     &       &  \\
\hspace{1em}Syphilis         & 2,336       & 2,603 \\
\hspace{1em}Inguinal Hernia  & 1,914       & 2,826 \\
\hspace{1em}Appendicitis     & 1,773       & 2,250 \\
\textbf{Frequent - Answers}     &       &  \\
\hspace{1em}Depression       & 1,140       & 11,715 \\
\hspace{1em}Anxiety          & 1,025       & 10,734 \\
\hspace{1em}Bipolar Disorder &   928       & 3,906 \\
\textbf{Frequent - Both}     &       &  \\
\hspace{1em}HIV and AIDS     & 5,664       & 6,162 \\
\hspace{1em}COVID-19         & 4,674       & 5,339 \\
\bottomrule
\end{tabular}

}
\caption{\label{tab:condition_distribution}
Most frequent conditions in the dataset, comprising the union of the five most frequent by questions and answers.}

\end{table}

Furthermore, the high standard deviations for token counts in both fields (37.5 for questions and 82.7 for answers) highlight a strong variability in linguistic depth. This variability is a key feature of the resource, demonstrating that the dataset captures a wide spectrum of interactions, from simple and direct inquiries to complex narrative-driven case descriptions, making it a robust resource for training models that can both understand diverse patient communication styles and generate appropriately detailed expert responses.

\subsection{Diversity and Coverage}

Beyond its scale, MedPT exhibits substantial thematic diversity, covering over 3,200 unique medical conditions. An analysis of the condition distribution reveals wide coverage across medical domains and highlights a key characteristic of the dataset. As shown in Table \ref{tab:condition_distribution}, the most frequent conditions among patient questions include infectious and clinical diseases such as HIV/AIDS, COVID-19, and Syphilis, while the conditions with the highest volume of professional answers are concentrated in mental health fields. This distinction underscores the dataset's unique value, capturing a broad spectrum of both urgent public health concerns from patients and areas of deep, sustained engagement from healthcare professionals.

Furthermore, Table \ref{tab:condition_distribution2} demonstrates that conditions with few questions often received a huge amount of answers, suggesting that even rare or highly specific issues tend to stimulate extensive professional interaction. This pattern indicates that, although these conditions represent a smaller portion of the dataset, they attract multiple perspectives and deeper discussions around the same underlying question.

\begin{table}[t]
\centering
\renewcommand{\arraystretch}{1.2} 

\adjustbox{width=1\columnwidth}{
\begin{tabular}{lcc}
\toprule
\textbf{Condition} & \textbf{Questions} & \textbf{Answers} \\
\midrule
Food desensitization         & 1       & 39       \\
Nightmares  & 1       & 38    \\
Changes After Death     & 2       & 75       \\
Relationship conflicts       & 3	   & 112       \\
Mutism          & 1	   & 35      \\
Existential distress     & 2       & 66      \\
Difficulty in decision-making         & 3      & 96      \\
Deep and prolonged sadness        & 3      & 95       \\
Shyness         & 10      & 311      \\
\bottomrule
\end{tabular}

}
\caption{\label{tab:condition_distribution2}
Conditions with few patient questions and a high number of answers.}

\end{table}

\begin{figure}[ht]
\centering
\includegraphics[width=0.48\textwidth]{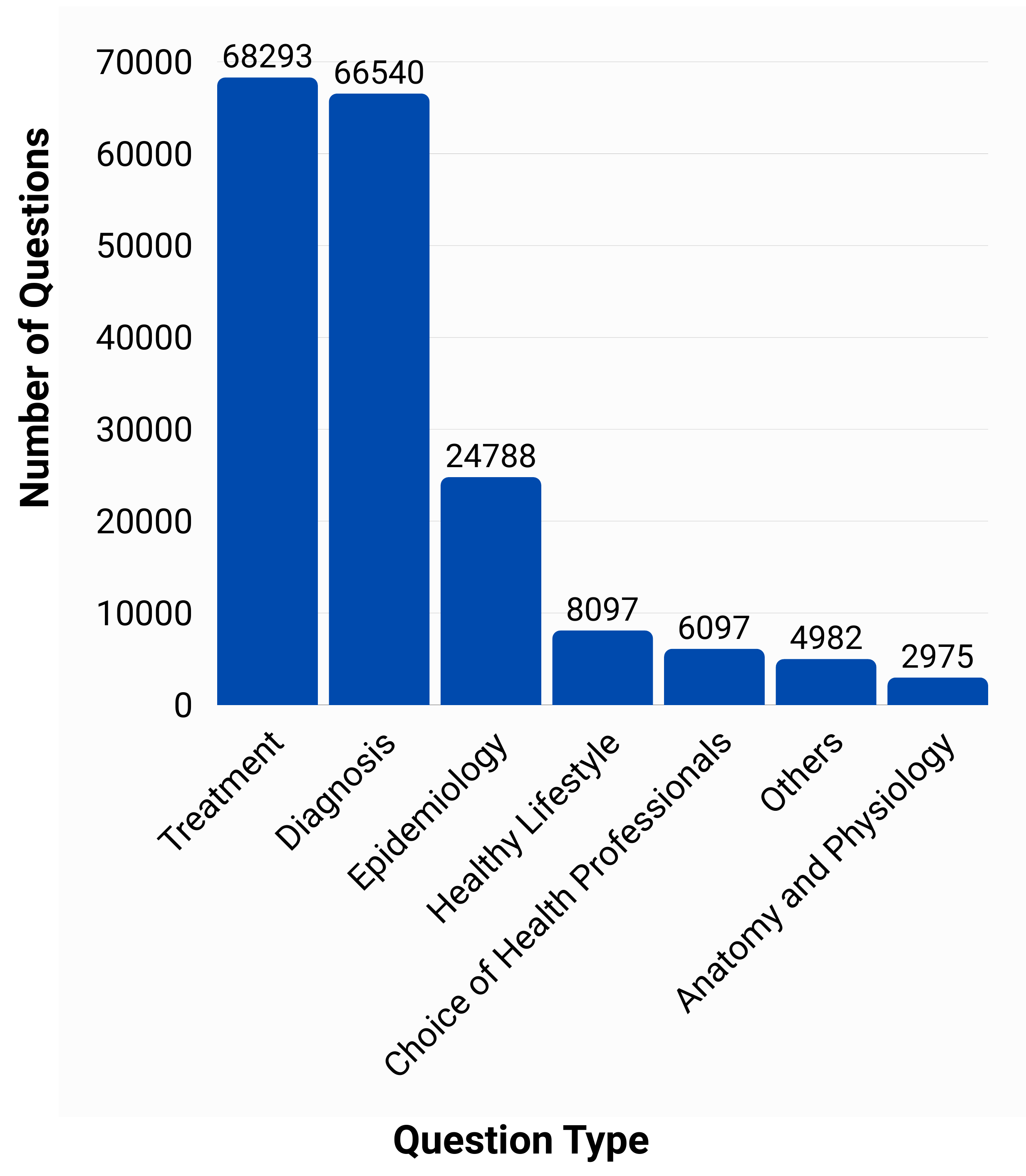}
\caption{Distribution of medical questions by type.}
\label{fig:question_type}
\end{figure}

To facilitate a finer-grained analysis of user intent, each question was enriched with a \textit{Question Type} label. The distribution of these categories, shown in Figure \ref{fig:question_type}, reveals that the dataset is dominated by action-oriented inquiries. Questions concerning \textit{Treatment} and \textit{Diagnosis} collectively account for nearly 70\% of the data, indicating a primary focus on actionable medical guidance. Concurrently, the presence of informational categories such as \textit{Epidemiology} and \textit{Healthy Lifestyle} confirms the dataset's thematic breadth. This diversity is also reflected in the specialties of the answer providers, including answers from professionals across 104 distinct specialties, which form over 900 unique specialization combinations due to multi-specialty declarations.

\begin{figure}[ht]
\centering
\includegraphics[width=0.48\textwidth]{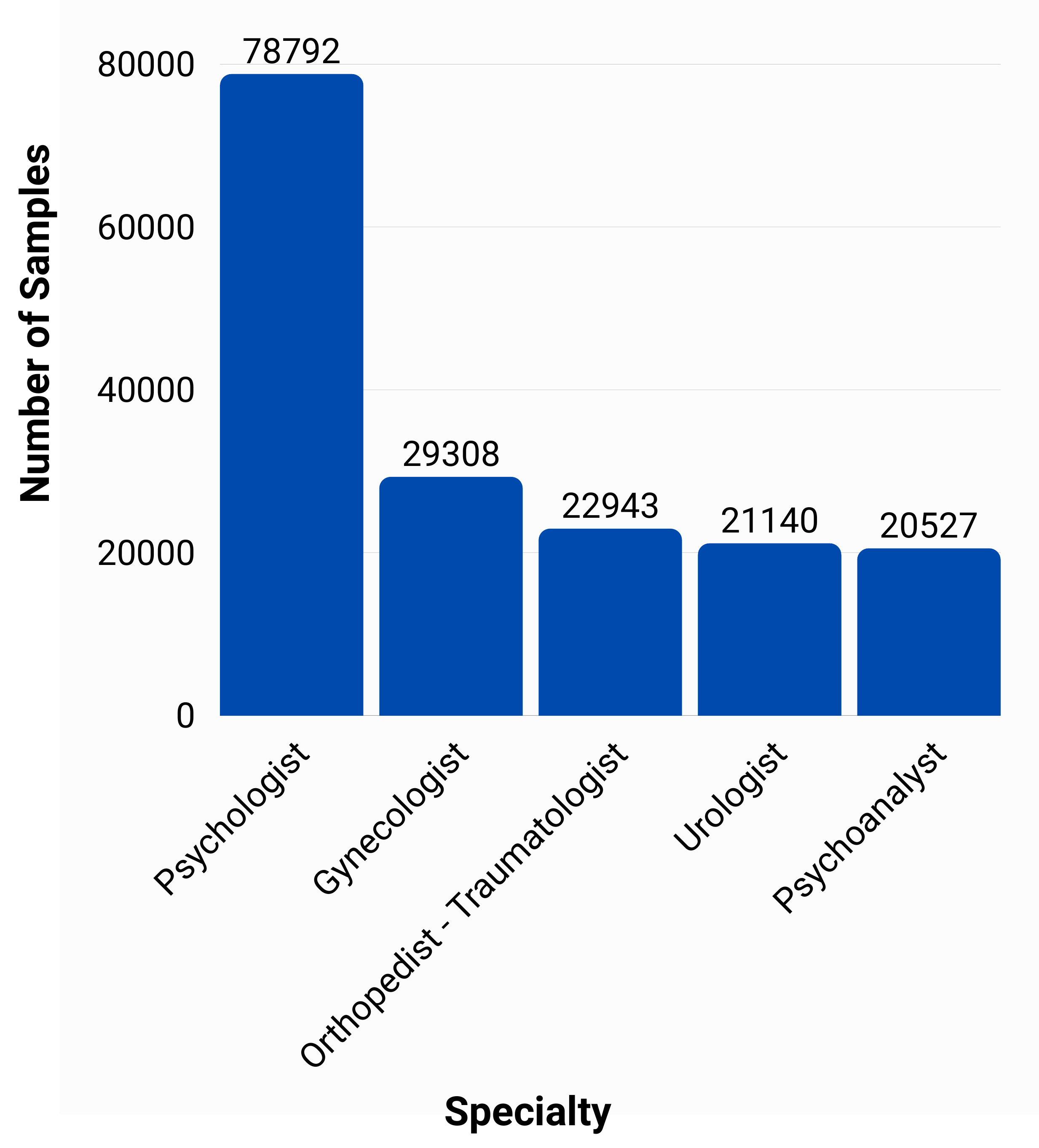}
\caption{Distribution of the top five most represented medical specialtys by number of answers.}
\label{fig:specialty_most}
\end{figure}

Furthermore, the data exhibits a long-tail distribution characteristic of real-world: high-frequency specialties such as \textit{Psychology}, \textit{Gynecology}, and \textit{Orthopedics} are well-represented (as shown in Fugure \ref{fig:specialty_most}), while a wide array of highly specialized fields, including \textit{Bariatric Surgery} and \textit{Pediatric Nutrology}, constitute the long tail with few number of examples (8 and 4 samples, respectively). This broad spectrum, spanning both common health concerns and niche medical domains, underscores the dataset's ecological validity as a resource for modeling authentic healthcare interactions, offering a broad and realistic coverage.

\section{Experiments}
To test MedPT dataset under downstream applications, we formulate a multi-class classification task designed to predict the appropriate medical specialty from a patient's question. This task simulates a critical real-world use case: the automated routing of patient inquiries to the correct clinical department. To evaluate model robustness against varying levels of label complexity, we create three experimental tiers based on the Top 5, Top 10, and Top 20 most frequent subspecialties in the dataset.

\subsection{Experimental Setup}
For all experiments, we use the Qwen3 1.7B \cite{yang2025qwen3} due to its balance between computational efficiency and performance, offering competitive results while maintaining a substantially lower training and inference cost compared to larger-scale models. We then benchmark its performance across three distinct methodological settings corresponding to the Top 5, Top 10, and Top 20 most frequent subspecialties. Each split as divided into an 80\% training set and a 20\% held-out test set. The evaluation settings are as follows:

\begin{enumerate}

\item \textbf{Zero-Shot:} The base model performs classification using only the task instructions and the patient's question, without any examples. This tests the model's intrinsic ability to generalize to the medical domain.

\item \textbf{Few-Shot (In-Context Learning):} The base model is prompted with instructions along with a small number of dynamically-sampled examples from the training set. We evaluate its in-context learning capability with 3 and 5 examples to measure performance with varying levels of contextual information.

\item \textbf{Fine-Tuning:} The model is fine-tuned on the 80\% training split of each respective sub-dataset. The training is conducted for 5 epochs using the AdamW optimizer with a learning rate of 2e-5 and a batch size of 32. This setting evaluates the dataset's effectiveness for specializing a model to the task.

\end{enumerate}

Performance across all settings is measured using standard classification metrics: precision, recall, and F1-score, calculated with macro averages to account for potential class imbalance.

\begin{table}[h]
\footnotesize
\centering
\resizebox{0.5\textwidth}{!}{%
\begin{tabular}{ll|ccc}
\hline
& \textbf{Strategy} & \textbf{Precision} & \textbf{Recall} & \textbf{F1-Score} \\
\hline

\multirow{4}{*}{\rotatebox{90}{\footnotesize \textbf{Top-5}}}
& Zero-Shot & 0.81 & 0.60 & 0.60 \\
& 3-Shot & 0.88 & 0.86 & 0.86 \\
& 5-Shot & 0.88 & 0.87 & 0.87 \\
& Fine-Tuned & \textbf{0.98} & \textbf{0.98} & \textbf{0.98} \\
\hline

\multirow{4}{*}{\rotatebox{90}{\footnotesize \textbf{Top-10}}}
& Zero-Shot & 0.79 & 0.55 & 0.54 \\
& 3-Shot & 0.84 & 0.76 & 0.78 \\
& 5-Shot & 0.85 & 0.77 & 0.79 \\
& Fine-Tuned & \textbf{0.96} & \textbf{0.96} & \textbf{0.96} \\
\hline

\multirow{4}{*}{\rotatebox{90}{\footnotesize \textbf{Top-20}}}

& Zero-Shot & 0.75 & 0.58 & 0.60 \\
& 3-Shot & 0.80 & 0.71 & 0.73 \\
& 5-Shot & 0.80 & 0.71 & 0.73 \\
& Fine-Tuned & \textbf{0.94} & \textbf{0.94} & \textbf{0.94} \\
\hline

\end{tabular}%
}
\caption{\label{tab:all_results} Results for classification.}

\end{table}

\subsection{Results}
The experimental results, summarized in Table \ref{tab:all_results}, reveal a clear and consistent performance hierarchy across all tested configurations. Fine-tuning the Qwen3-1.7B model on the MedPT dataset yields the highest performance by a significant margin, establishing a strong benchmark and demonstrating the dataset's value for model specialization. As expected, performance for all methods degrades systematically as the task difficulty increases from 5 to 20 classes. Nonetheless, the fine-tuned model maintains an exceptional 0.94 F1-score on the most challenging 20-class task, underscoring the quality of the learned representations.

The efficacy of in-context learning (ICL) is also clearly demonstrated. In the 20-class setup, for example, the zero-shot baseline achieves a 0.60 F1-score, indicating the task's inherent difficulty. However, by providing just three examples (3-shot), the score jumps to 0.73. This substantial improvement highlights the model's ability to quickly leverage the contextual patterns within MedPT to enhance its predictive accuracy without full fine-tuning.

A key finding from the ICL experiments is a clear point of diminishing returns. The performance gain from zero-shot to 3-shot is substantial across all settings. However, increasing the examples from 3-shot to 5-shot provides only a marginal benefit (e.g., 0.73 F1-score for both in the 20-class setup), suggesting that a 3-shot prompt may represent the optimal balance of performance and efficiency for this task.

\section{Discussion}
The experimental results confirm the high utility of MedPT for developing specialized medical language models. The exceptional performance of the fine-tuned model, achieving an F1-score of 0.94 on the medical specialty classification task involving 20 specialties, directly validates the dataset's quality. This demonstrates that the linguistic patterns and class labels within MedPT are consistent and highly learnable, providing a robust foundation for domain-specific classifiers. Furthermore, our findings offer practical implications for deployment: the substantial performance gains from in-context learning confirm MedPT's value for resource-light applications, while the observed diminishing returns between 3-shot and 5-shot prompts suggest an optimal efficiency-performance trade-off.

A deeper qualitative analysis of the fine-tuned model's errors revealed that misclassifications were not random but systematic, occurring between semantically and clinically related subspecialties. We observed notable confusion between conditions with high comorbidity (e.g., \textit{Anxiety} and \textit{Depression}), anatomical proximity (e.g., \textit{Umbilical Hernia} and \textit{Inguinal Hernia}), or overlapping symptoms (e.g., \textit{Gastritis} and \textit{Gastroesophageal Reflux}). This pattern indicates that the model is capturing a meaningful semantic space of medical concepts, where its errors reflect genuine clinical ambiguity, underscoring MedPT's potential to support future research into more complex tasks, such as fine-grained medical entity disambiguation.

\section{Conclusion}
In this work, we introduced MedPT, the first large-scale, real-world dataset of patient-doctor interactions in Brazilian Portuguese, containing over 384,000 curated question-answer pairs. This resource was developed to address the critical gap of culturally and clinically relevant data for a major but under-resourced language, capturing authentic user inquiries that include endemic diseases and local health concerns often absent in translated datasets. Our contribution extends beyond collection to the rigorous, multi-stage curation protocol itself. We applied a hybrid quantitative-qualitative analysis to every feature, using statistical distributions to guide deep manual inspection. This meticulous process, coupled with an LLM-driven annotation of user intent, ensured the dataset's high signal-to-noise ratio and structural richness.

This high utility was then confirmed through multi-faceted validation. In downstream classification experiments, a 1.7B-parameter model fine-tuned on MedPT achieved an F1-score of 0.94 on the medical specialty classification task across 20 specialties, compared to 0.60 for the base model. This, combined with strong few-shot ICL results, proves the dataset's practical value. By publicly releasing MedPT, we provide a foundational resource to the research community. We aim to support the development of more equitable, accurate, and safe medical NLP technologies, enabling a new generation of models that can understand the specific clinical and linguistic nuances of the Portuguese-speaking world.

\section{Limitations}
We acknowledge several limitations in this work that also present clear avenues for future research. First, while our experimental validation focused exclusively on a multi-class classification task demonstrates the dataset's utility for a crucial application like inquiry routing, its performance on generative tasks, such as medical question-answering and patient-doctor dialogue, remains to be explored. Second, our deduplication process was based on removing exact-match samples instead of more sophisticated near-duplication detection techniques. Finally, regarding data quality, although all answers originate from verified physicians, a secondary, independent clinical review of the responses could further reinforce the dataset's reliability and ensure the clinical accuracy of the provided information.

\section{Ethical Considerations}
The MedPT dataset and any models developed using it are intended to function as assistive tools to support, not replace, the judgment of qualified healthcare professionals. We emphasize that AI systems trained on this data should not be used as a substitute for professional medical consultation, diagnosis, or treatment. Relying solely on LLM-generated content for medical decisions carries significant risks, as models may produce outputs that are plausible but clinically inaccurate. The primary ethical application of this work is to augment clinical workflows, for tasks such as inquiry routing, information retrieval, preliminary symptom summarization, while ensuring that the final diagnostic and therapeutic decisions remain the exclusive responsibility of a human medical expert.








\section{Acknowledgments}
This work has been fully/partially funded by the project Research and Development of Algorithms for Construction of Digital Human Technological Components supported by Advanced Knowledge Center in Immersive Technologies (AKCIT), with financial resources from the PPI IoT of the MCTI grant number 057/2023, signed with EMBRAPII.

\section{Bibliographical References}\label{sec:reference}

\bibliographystyle{lrec2026-natbib}
\bibliography{lrec2026-example}

\bibliographystylelanguageresource{lrec2026-natbib}
\bibliographylanguageresource{languageresource}

\appendix
\section{Question-Type Annotation Prompt}
\label{app:prompt}

The following prompt (originally written in Brazilian Portuguese) was used to instruct \texttt{gpt-oss-120b} to classify each patient question into one of seven predefined semantic categories, as described in Section~\ref{sec:annotation}.

\begin{tcolorbox}[colback=gray!10, colframe=gray!50, title=Prompt, fonttitle=\bfseries, coltitle=black, fontupper=\small]
Act as a classifier specialized in medicine, with deep knowledge of all types of medical questions. Your task is to classify medical questions based solely on the content of the question.

\textbf{Instructions:} 

(1) Classify the question into only one of the types listed below. 

(2) Never classify into more than one type.

(3) If it is not possible to classify with confidence, classify as ``Others''. 

(4) Your answer must contain only the TYPE of the question. Nothing else. 

(5) Do not invent types that are not listed.

\textbf{Available types:}

Diagnosis, Treatment, Anatomy and Physiology, Epidemiology, Healthy Lifestyle, Choosing Healthcare Professionals, Others.

\textbf{Data to be classified:}\\
{-}{-} Start of data {-}{-}\\
Question: \{question\}\\
{-}{-} End of data {-}{-}
\end{tcolorbox}

\end{document}